\documentclass[letterpaper, 10 pt, conference]{ieeeconf}  

\IEEEoverridecommandlockouts                              

\usepackage{graphicx}
\usepackage{gensymb}
\usepackage{algorithm, algpseudocode}
\usepackage{subfigure}
\usepackage{url}
\usepackage{subfloat}

\title{\LARGE \bf
Towards Full Automated Drive in Urban Environments:\\A Demonstration in GoMentum Station, California
}

\author{Akansel Cosgun$^{1}$, Lichao Ma$^{1}$, Jimmy Chiu$^{1}$, Jiawei Huang$^{1}$, Mahmut Demir$^{1}$, Alexandre Miranda A\~non$^{1}$, \\Thang Lian$^{1}$, Hasan Tafish$^{2}$, Samir Al-Stouhi$^{2}$
\thanks{$^{1}$Authors are with Honda Research Institute, Mountain View, CA, USA {\tt\small ACosgun@hra.com}, {\tt\small MMa@hra.com}, {\tt\small JChiu@hra.com}, {\tt\small JHuang@hra.com}, {\tt\small MDemir@hra.com}, {\tt\small AMiranda@hra.com}, {\tt\small TLian@hra.com}}
\thanks{$^{2}$Authors are with Honda Research and Development America, Detroit, MI, USA {\tt\small HTafish@hra.com}, {\tt\small SAlStouhi@hra.com}}
}

\begin{document}

\maketitle

\thispagestyle{empty}
\pagestyle{empty}

\begin{abstract}

Each year, millions of motor vehicle traffic accidents all over the world cause a large number of fatalities, injuries and significant material loss. Automated Driving (AD) has potential to drastically reduce such accidents. In this work, we focus on the technical challenges that arise from AD in urban environments. We present the overall architecture of an AD system and describe in detail the perception and planning modules. The AD system, built on a modified Acura RLX, was demonstrated in a course in GoMentum Station in California. We demonstrated autonomous handling of 4 scenarios: traffic lights, cross-traffic at intersections, construction zones and pedestrians. The AD vehicle displayed safe behavior and performed consistently in repeated demonstrations with slight variations in conditions. Overall, we completed 44 runs, encompassing 110km of automated driving with only 3 cases where the driver intervened the control of the vehicle, mostly due to error in GPS positioning. Our demonstration showed that robust and consistent behavior in urban scenarios is possible, yet more investigation is necessary for full scale roll-out on public roads.

\end{abstract}

\section{Introduction}
\label{sec:intro}

Traffic accidents pose a great threat on public safety and have a significant negative effect on the economy. An estimated 35 thousand people lost their lives due to motor vehicle traffic crashes in 2015 in the United States alone. Moreover, there were more than 6 million accidents that resulted either in personal injuries and/or material damage, according to National Highway Traffic Safety Administration (NHTSA) \cite{nhtsa}. The majority of the reported crashes is a direct result of human error.

Automated Drive (AD) has great potential to drastically reduce the number of accidents. Over the past two decades, AD has garnered great interest. After the successes at the DARPA Grand and Urban Challenges \cite{urmson2008autonomous,montemerlo2008junior}, both academia and industry started showing more interest in AD. Moreover, recent developments in artificial intelligence and machine learning has made large-scale full AD a real possibility, with significant investment from automakers \cite{ziegler2014making} and tech companies such as Google and Uber. As the costs of sensors and electric cars go down, AD will be more and more economically feasible. In addition to saving lives, AD will also increase the productivity of the workforce as people will dedicate less time operating vehicles for transportation.

\begin{figure}[ht!]
  \centering
  \includegraphics[width=1.0\linewidth]{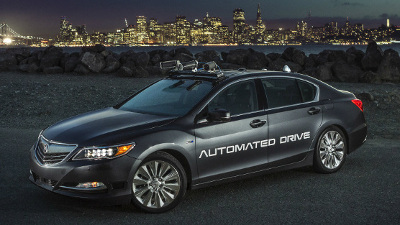}
\vspace{-0.5cm}
  \caption{A 2016 Acura RLX Sport Hybrid SH-AWD, specifically modified for Automated Driving, was used for the demonstrations}
  \label{acura}
\vspace{-0.6cm}
\end{figure}

There are legal, social and technical challenges in making AD a reality. In this paper, we address the technical aspect of our efforts towards full AD. In terms of technical requirements, AD falls into two domains: highway and urban. On a highway, possible situations that can occur are limited, therefore highway AD is relatively easier to handle. Urban AD, on the other hand, is typically harder to tackle because of the complexity that arises in cities due to variability of environments, agents (pedestrians, vehicles, bicycles), traffic rules and interactions between these.

In this work, we focus on the technical challenges that arise from realizing a system capable of demonstrating full AD in scenarios that can arise in urban environments. We first talk about the vehicle in Section \ref{sec:vehicle}, a 2016 Acura RLX augmented with a sensor suite, hardware configuration and power system suited for AD. Then we describe the system architecture and describe how the hardware, perception and decision making modules interact in Section \ref{sec:system}. In Section \ref{sec:perception}, we present perception modules: pedestrian, vehicle and obstacle detection as well as localization and Vehicle-to-Anything (V2X) communication. In Section \ref{sec:planning}, we present planning modules: behavior, and trajectory planning. In Section \ref{sec:demo}, we describe the conditions of the demonstration and how the AD vehicle tackled scenarios that are commonly encountered in urban environments. Specifically, we demonstrate AD capabilities for 4 scenarios: handling traffic lights, cross-traffic at intersections, pedestrians and construction zones. The demonstrations were conducted in GoMentum Station in Concord, CA with some variability in agents' behavior and were a success in terms of safety and robustness. Finally, in Section \ref{sec:conclusion}, we sum up conclusions and discuss future direction. 
\section{Vehicle}
\label{sec:vehicle}

The 2016 Acura RLX Sport Hybrid SH-AWD, seen in Figure \ref{acura}, is modified for automated driving, to be driven by a computer which generates the trajectory and motion of the vehicle in real time based on the information gathered from different sensors installed on the vehicle.

	The front camera is used for pedestrian detection which is mounted on the rack on top of the roof located approximately above the rear view mirror. Its resolution, visible azimuth Field of View (FOV) and maximum range are $1920\times 1200$, $80\degree$ and $80m$, respectively.
	
	Three LiDARs provide the point cloud in front of the vehicle which is being used for obstacle and pedestrian detection. The front LiDAR is installed in the middle of the vehicle in front and the side LiDARs are installed at $75\degree$ on each side of the vehicle behind the front bumper cover. The LiDAR azimuth FOV is $110\degree$ and the detection range is $100m$.
	For lateral localization where GPS is not reliable, a camera which processes lane/vehicle detection, is mounted behind the windshield near the center of the vehicle. The sensor FOV is $40\degree\times 30\degree$ (WxH) and has detection range of $80m$.
	Two side-facing ($90\degree$) radars are mounted inside the front bumper, one on each side of the vehicle, used for detecting oncoming traffic at the intersections. The radar’s long-range detection is $175m$ with azimuth FOV of $20\degree$ and mid-range detection is $60m$ with azimuth FOV of $90\degree$.
	
	RTK GPS provides real time positioning of the vehicle and uses INS technology and measures the vehicle motion, position and orientation at a rate of $250Hz$. The velocity, position and orientation estimation errors are within $0.05km/h$, $0.01m$ and $0.1\degree$, respectively, in RTK mode. The unit uses two antennas mounted one on the moon-roof and the other near the back windshield.
	
	The camera and LiDAR data is processed by the PC installed in the back seat of the vehicle which is powered by two auxiliary batteries installed behind the driver seat. The auxiliary batteries are charged when the vehicle battery is fully charged through a battery charger circuit which is embedded in system powering unit. The rest of the units including the planner computer are installed in the trunk and powered by 12V converted from high voltage hybrid battery.
	
	The vehicle is also equipped with a V2X transceiver to receive messages from, and transmit to other V2X-equipped vehicles and pedestrians.
\section{System Overview}
\label{sec:system}

\begin{figure}[ht!]
  \centering
  \vspace{4pt}
  \includegraphics[width=1.0\linewidth]{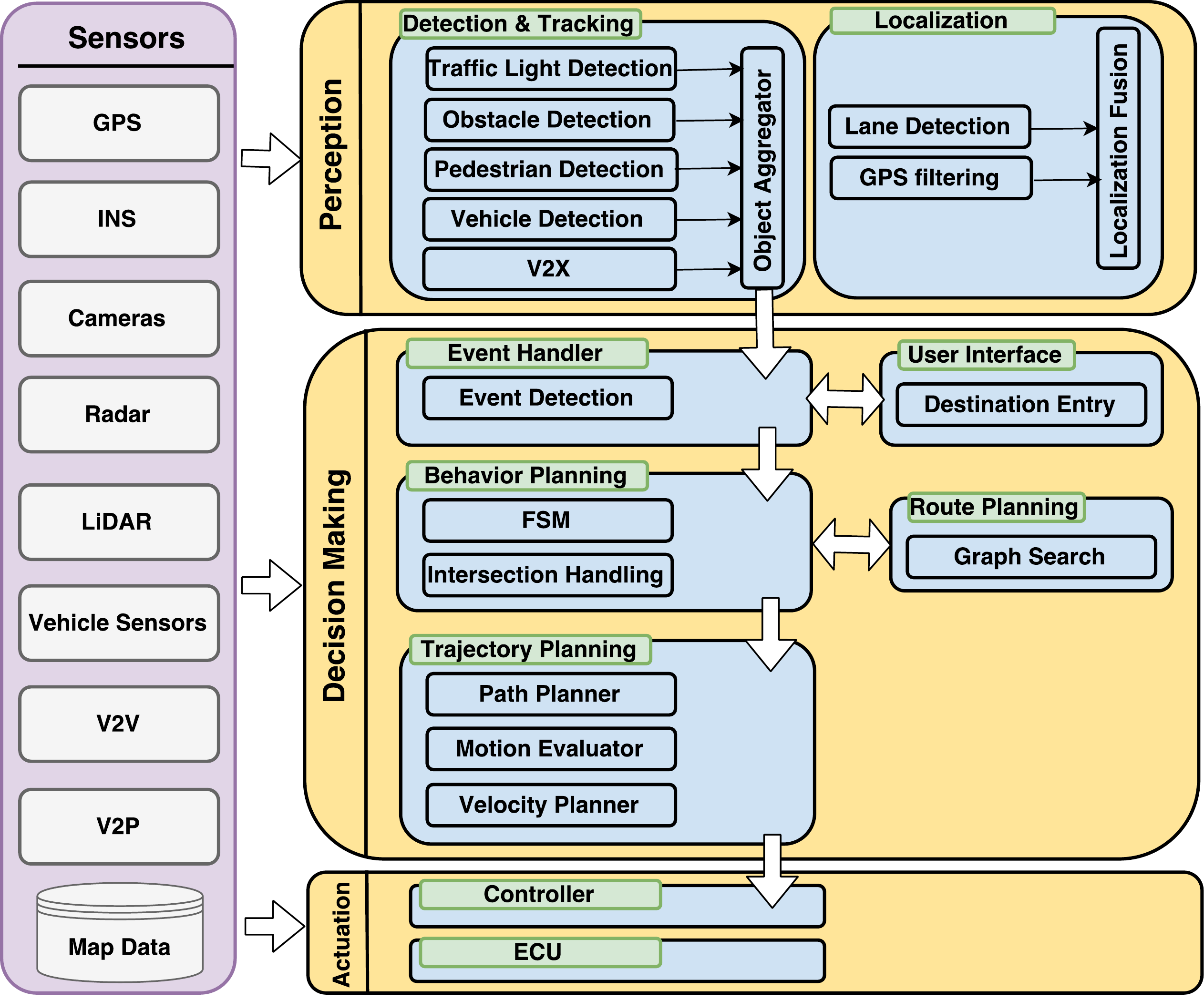}  
  \caption{System Overview of the AD vehicle}
  \vspace{-0.7cm}
  \label{sys_diagram}
\end{figure}

The AD system comprises several complex and interacting modules. These modules can be grouped into four, as seen in Figure \ref{sys_diagram}: sensor, perception, decision making and actuation. 

Sensor modules provide raw data to all other modules that request it, using sensor-specific firmware. These modules include GPS/INS, cameras, radar, LiDAR, V2X and other vehicle sensors. Map data is organized and served to any process that requires information from the map.

Perception modules include localization and detection/tracking modules. Road lanes are detected, GPS data is pre-filtered and the data is fused by localization modules. Several processes are dedicated to detection and tracking of objects. These include traffic light, pedestrian and vehicle detection using predominantly computer vision. Obstacle detection module, using LiDAR, detects any obstacles that don't fit to a model (i.e. vehicle) and should be avoided by the AD vehicle. V2X module provides data from, and to other V2X-equipped vehicles and pedestrians. If available, Vehicle-to-Pedestrian (V2P) and Vehicle-to-Vehicle (V2V) modules relax the line-of-sight requirement of on-board sensing and helps the AD vehicle 'see' and reason about occluded objects. \textbf{Object Aggregator} module takes in all the detection and tracking results, fuses the data, applies temporal and geo-filtering, and sends a coherent output to the \textbf{Event Handler}.

Decision making modules consist of route, behavior and trajectory planning. We use an event-based system where object tracks reported by the Object Aggregator trigger events. Event Handler module receives information from Object Aggregator, and raises event(s) if the object data is consistent with a pre-defined set number of rules. For example, traffic light state changes or the appearance of pedestrians would raise events. User requests are also handled as events, for example when a new destination or a stop request is entered. The events are received, queued and run through a Finite State Machine (FSM), which in turn triggers a state change if the inputs conform to the state transition table. Behavior planning is also involved with handling pedestrians, other vehicles and intersections. The FSM can any time query a route given a start location and destination. Behavior planning layer determines a sub-goal point along the route and passes it to trajectory planner layer. Trajectory planning module first plans a nominal path toward the sub-goal, evaluates primary and alternative collision-free paths and plans for the velocity. 

The reference trajectory is passed to the actuation layer, where the controller determines the steering and gas/braking so the deviation from the reference trajectory is minimized.
\section{Perception}
\label{sec:perception}

In this section, we present modules related to perceiving the environment. We first describe our localization approach that corrects lateral localization using lane markings. We then present our vision and LiDAR-based pedestrian detection, followed by radar-based vehicle tracking algorithm. We then describe our obstacle detection approach for avoidance. Finally, we touch on our usage of V2P sensing.

\vspace{-0.2cm}

\subsection{Localization}

This section describes the localization algorithm used to improve the \textit{lateral} localization within the ego lane. The fusion of L1-Global Positional Systems (GPS) with either Inertial Measurement Units (IMU) or other proprioceptive sensors like an odomoter using an Extended Kalman Filter (EKF) or Unscented Kalman Filter (UKF) is known to produce a smoother and more reliable estimation of the pose of the vehicle \cite{cui_ekf_urban} \cite{vdw_ukf} than simple GPS. However, the accuracy is often not enough for AD, especially for urban environments where GPS outages (trees, tunnels), multipath errors or biased errors (ionospheric delay) are common. In order to tackle this problem, researchers \cite{tao_localization} \cite{gruyer_localization} \cite{jo_localization} have shown that using digital maps of the lane markings made \textit{a priori} in combination with a camera facing the road can improve the lateral position estimation of the vehicle. Our approach is similar to these works.


Our system utilizes 3 main inputs:
\begin{itemize}
\item GPS/INS system that provides global position estimation. This system integrates the L1-GPS, IMU and odometer sensor using an EKF.
\item A lane detection algorithm that uses a monocular camera to detect the ego lane and fit a third degree polynomial \(a + b x + c x^2 + d x^3\) to the lane marking. This system is calibrated so that the distances provided are already in the ego car frame. In our case, we use the (a) distance and (b) the heading angle for the left and right lanes. The lane detector also provides the type of lane marking (dashed, solid, road edge, etc.) and the quality of the estimation.
\item Map data with highly accurate globally positioned lane markings (acquired using RTK-GPS) along with the type of the lane marking.
\end{itemize}

Pseudocode of the lateral localization algorithm is shown in Algorithm \ref{alg:alg_loc}. Whenever GPS/INS data is received, it is sent directly to the fusion algorithm. Otherwise, if an input from the vision system is received, first the map matching procedure is applied, followed by a rejection strategy to identify the closest lane segments on the digital map to the left and the right of the car. The distances from the center of the car to the selected line segments are then computed and sent to the fusion algorithm. During initialization, or when there is no input from the vision system, we employ a special case, where a one-step particle filter is applied to re-lock the position. 

\vspace{-0.2cm}

\subsection{Pedestrian Detection \& Tracking}

The goal of the pedestrian detector is to accurately detect pedestrians' position and velocity in 3D. It consists of three stages. The first stage is pure vision-based pedestrian detection. We first extract Histogram of Gradients (HOG) features from the image at different scales. The extracted features are then classified by a linear Support Vector Machine (SVM), followed by a Radial Basis Function (RBF). Using RBF is crucial because it significantly increases the discriminatory power of the classifier. Both classifiers are trained on 5000 positive samples and iteratively trained with hard negative samples. The ouput of this stage is a set of bounding boxes and corresponding scores indicating the likelihood of containing a pedestrian. To reduce false-positive rate, we discard any bounding box which scores below an empirically derived threshold of 0.5.

\setlength{\floatsep}{0.1cm}

\begin{subfigures}
\begin{figure}
\vspace{2mm}
\end{figure}
\begin{algorithm}[t]
\caption{Lateral localization algorithm} \label{alg:alg_loc}
\begin{algorithmic}[1]  
\State Initialize ${pos_{car}}$ based on GPS
\Loop
\If{${input}$ = GPS/INS}
\State ${FusionAlgorithm} \ \ \textbf{return} \ {pos_{car}}$
\ElsIf{${input}$ = Vision System}
\If{Initialization \textbf{or} First input from Camera}
\ForAll {${Particles}$}
\State ${MapMatching \Rightarrow Lines}$
\State ${ComputeDistances \Rightarrow D_{Left}, D_{Right}}$
\EndFor
\State Select best particle based on likelihood
\State ${FusionAlgorithm} \ \ \textbf{return} \ {pos_{car}}$
\EndIf
\State ${MapMatching \Rightarrow Lines}$
\State ${ComputeDistances \Rightarrow D_{Left}, D_{Right}}$
\State ${FusionAlgorithm} \ \ \textbf{return} \ {pos_{car}}$
\EndIf
\EndLoop
\end{algorithmic}
\caption{Lateral localization algorithm} \label{alg:alg_loc}

\end{algorithm}
\end{subfigures}

The second stage is LiDAR association to estimate distance \cite{heisele2017system}. Using the extrinsic calibration between the LiDAR and the camera, we can project all the LiDAR points located in front of the camera onto the image. Then for each pedestrian bounding box detected in the first stage, we find all LiDAR points located inside the box. Among these points, we pick the one closest to the ego vehicle and assign its coordinates to the bounding box. The reason is that only the closest LiDAR point actually belongs to the pedestrian. After each bounding box's coordinates are assigned, we perform several heuristic-based filtering steps - each bounding box must pass all tests or be removed. For example, a bounding box must not be entirely above the horizon line, the box must have typical aspect ratio of a person, the box must have normal sizes at the given distance and the box must have at least one LiDAR point inside. The remaining bounding boxes are passed to the third stage. After this step, a pedestrian will be represented as a 3D point.

The third and last stage is the linear tracker, that estimates pedestrian velocity. We perform tracking in a fixed global frame, by transforming all vehicle-relative coordinates into a fixed global frame. After the transformation, we use a linear motion model to track the pedestrian. The algorithm works as follows: first we cluster all the points. If a cluster is close enough to an existing tracker's latest position, it is considered the latest observation for that tracker and be appended to that tracker. The tracker's trajectory is then updated to the least-square fit linear curve of all the assigned observations. If a cluster does not lie close to any existing tracker, we create a new tracker for it. In addition, we delete any tracker that has no update for 10 consecutive frames, deeming that the object has disappeared from view and that the track is lost.

The final output of the pedestrian detection module is the current position and velocity of all tracked pedestrians. These are eventually consumed by the event handler to raise a pedestrian event.

\vspace{-0.2cm}

\subsection{Vehicle Detection \& Tracking}

One challenging task for AD is to detect cross-traffic in intersections. For reliable handling of intersections, we'd like to detect cross traffic vehicles from $150m$ away. At this distance, camera is not effective because the target car only shows in a small number of pixels. Similarly, number of hits from LiDAR is too few for classification. Therefore, we use radar-based cross-traffic detection.

We utilize mid-range radars which combines two operating modes: long-range narrow-angle FOV and mid-range wide-angle FOV. The device alternates between the two modes at a frequency of $40Hz$. By assembling the two consecutive scans, we create a full scan at a frequency of $20Hz$. In the long-range mode, the radar can detect up to $175m$ which is sufficient for our purposes.

The radar points are first filtered by their velocity, position and signal strength. The filtered points are then clustered. Depending on their position, each cluster either gets assigned to an existing tracker or generates a new tracker. We use particle filtering as our backend tracking algorithm. We use the particle filter state and measurement models similar to the one presented in \cite{knill2016direct}.

Each tracked vehicle is represented by a state vector $\zeta_k = [x_k, y_k, v_k, \theta_k, \omega_k]$. $(x_k, y_k)$ is the vehicle position at $t_k$, $v_k$ is the vehicle velocity, $\theta_k$ is the vehicle heading angle and $\omega_k$ is the vehicle yaw rate. 


\vspace{-0.2cm}

\subsection{Obstacle Detection}

We detect static obstacles using several LiDAR sensors. LiDAR provide a sparse representation of the environment, especially at distance, therefore LiDAR-based obstacle detection is a challenging task. We focus on detection of construction zone obstacles, such as cones, however our method can be extended for a wider set of obstacles. Our approach consists of several consecutive layers of filtering and clustering operations.

In the first step, LiDAR scans are cropped as to get points in the front FOV only. Then, the ground plane is extracted method to filter out the points reflected from the road surface. The remaining points are clustered based on an euclidean distance based heuristic. Each cluster is then analyzed according to its attributes such as cluster size, relative distance from the ego vehicle, cluster bounding box area and height. Clusters with following criteria are selected as obstacle candidates: cluster center must be close to the road surface and bounding box area and height should be greater than predefined minimum obstacle dimensions. However, these criteria need to be relaxed at certain conditions due to the sparsity of sensor data: If the distance of cluster center to the ego vehicle is greater than a certain threshold, it is selected as  an obstacle candidate regardless of its bounding box area and height. Moreover, if the cluster center is too close to the ego car, vertical angle between LiDAR rays limits the maximum perceivable height. Therefore, minimum bounding box height limit is lowered to two thirds of maximum vertical distance between LiDAR rays at that distance. In our experience, these two modifications improved our detection rate considerably for both far and near obstacles. In the final step, spatio-temporal filtering is applied to simultaneously determine persistent obstacles among a set of obstacle candidates and filter out noisy observations. In order to implement this filter, obstacle candidates are first stored in a dynamic set of clusters where each cluster is updated periodically so they contain only the recent obstacle candidate. Clusters that contain more than a certain number of obstacle candidates are deemed obstacles and respective 2D bounding boxes are sent to the Motion Evaluator, to be factored in decision making.

\vspace{-0.2cm}

\subsection{Vehicle-to-Anything (V2X) as a supplementary sensor}


V2X can be an useful addition to AD perception suite. It is inexpensive, long range (up to $300 m$), and can see through obstructions. State-of-the-art V2X technology uses 802.11p physical layer, thus, it is easy to integrate it into any WiFi-enabled device. Currently, AD problem is addressed using on-board sensing and decision making. V2X technology has great potential for future AD applications, such as collective perception for overcoming the on-board sensor limitations \cite{dao2007markov}, infrastructure sensors for submitting traffic information and road conditions, as well as connected traffic lights for smooth handling of intersections.



In this work, we investigated V2X potential to enhance the effectiveness of pedestrian detection. For example, consider a pedestrian that is occluded by a parked truck. In this case, the cameras might detect the pedestrian too late, causing the vehicle to hard-brake or collide with the pedestrian. Therefore, we used a smart phone with a modified WiFi chip, operating in the 5.9 GHz ITS band, to send the pedestrian’s GPS location to the AD system. The phone periodically broadcasts Basic Safety Messages (BSM), containing the object's location, speed, acceleration, and type (i.e. pedestrian, vehicle, etc.).

The V2X transceiver in the AD vehicle parses received BSMs, determines if the message belongs to a pedestrian or a vehicle, transforms the object’s geodetic coordinates to the global frame and submits the results to the Object Aggregator. The detections are sent at $10 Hz$ but the rate is not guaranteed due to packet loss. Therefore, we use a Kalman Filter with constant acceleration motion model to smooth out the V2X object position. Tracking is re-initialized if no BSMs are received for more than a second. 

\section{Motion Planning and Control}
\label{sec:planning}

In this section, we present modules related to decision making. We first describe our state machine, followed by route planner, which is based on a graph search algorithm over nodes acquired from the map. We then talk about behavior planning, specifically how intersections and pedestrians are handled and how all inputs are managed by the state machine. We then present our trajectory generation algorithm and conclude with a brief description of the controller.

\subsection{State Machine}

\begin{figure}[ht!]
  \centering
  \vspace{4pt}
  \includegraphics[width=1.0\linewidth]{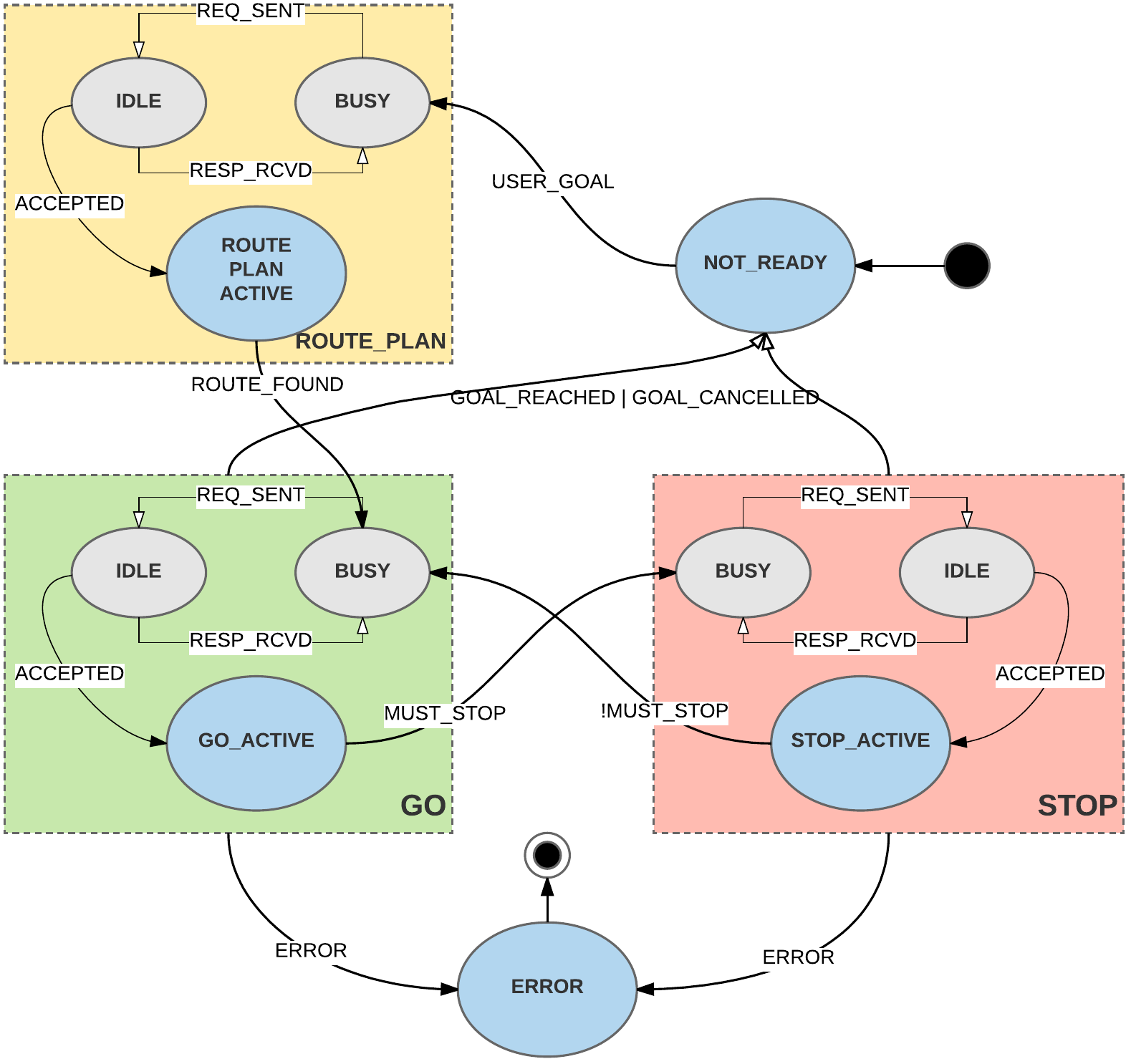}  
  \caption{We use a Hierarchical State Machine for high-level reasoning}
  \label{state_machine}
  \vspace{-0.2cm}
\end{figure}

We use a hierarchical state machine for level reasoning, from events that are generated by the Event Handler. A visualization of the state machine is shown in Figure \ref{state_machine}. The system has five main states:

\begin{itemize}
\item \textbf{NOT\_READY: }State where there is no destination goal
\item \textbf{ROUTE\_PLAN: }State when the route is being planned
\item \textbf{GO: }State when there is no requirement for stopping
\item \textbf{STOP: }State when the car has to come to a stop
\item \textbf{ERROR: }State where the vehicle quits AD mode due to error
\end{itemize}

The system starts in \textbf{NOT\_READY} state. When a user enters a destination, the state becomes \textbf{ROUTE\_PLAN}. If a plan is found, \textbf{GO} state is activated. In \textbf{GO} state, the Trajectory Planner is instructed to go in the current lane with suitable speed. Every time the state machine receives an event, an assessment is made to determine if it raises the \textbf{MUST\_STOP} flag. \textbf{PEDESTRIAN}, \textbf{TFL\_RED} or \textbf{INT} (intersection) events all trigger the switch to \textbf{STOP}. The latest events from these three event types are stored in memory, and sent to the go server  as requirements. For example, when the AD vehicle is within some distance to an intersection, \textbf{INT} event raised, which triggers switching to \textbf{STOP} state. While in that state, if a pedestrian shows up, it is a added to the requirements that should be resolved before switching back to \textbf{GO}. The requirements are satisfied with \textbf{PED\_CLEAR}, \textbf{TFL\_GREEN} and \textbf{INT\_OK}, respectively, for requirements above. When the goal is reached or when user cancels the goal, state switches to \textbf{NOT\_READY}. Errors that are not handled by the state machine cause the vehicle to go to \textbf{ERROR} state, and leads to disengagement from AD.

In practice, the algorithms for the state machine, route planner, going and stopping run in parallel processes. Suppose that the state machine got a user destination request and wants to switch to \textbf{NOT\_READY} from \textbf{GO}, but the route planner crashes or cannot find a solution. In this case, the state machine should stay in \textbf{NOT\_READY} instead of prematurely switching to \textbf{GO} as the go server wouldn't have received the global path to work with. Because of the parallel architecture of our state machine, we implemented a hand-shaking protocol where the a request is sent to the target state, and the state is changed only when a positive response is received. We use a communication sub-state machine that keeps track of whether a request is sent, and a response is received. \textbf{BUSY} sub-state means a request for connection has been sent for that state and a response is expected. Sub-state machines goes into \textbf{IDLE} when a response is received or a timeout occurs. It is also possible that the target state process is not ready and that it declines the requests. These situations raise either failure or timeout events, which gives the state machine feedback for error recovery. The processes send periodic messages that also serves as a heartbeat message, so that the state machine monitors the health of the processes that are supposed to be running.

\subsection{Route Planning}

We use a map that includes different layers of map information, such as lane center lines, road edges, cross walk zones, stop lines, etc. with high positional accuracy. Depending on the application, information from different layers are utilized. For example, the lane center lines are mainly used for navigating the vehicle. The road edges are used to filter out obstacles off the road. The stop line is used to trigger the \textbf{INT} event. After the map is loaded into the memory, we create a directed graph which consists of lane center lines. Then, we use A* search algorithm to search for a feasible route for a given start and goal position. After the route is calculated, the route planning module publishes this global path to all other modules. The global path has index and 3D coordinate on each point, as well as the legal speed limit. Both index and 3D coordinate are used to locate the vehicle in the map and trigger corresponding events.

\subsection{Behavior Planning}

This section describes how the AD vehicle behaves in the \textbf{STOP} state. When the vehicle goes into the \textbf{STOP} state, or is already in that state, several requirements can be active, related to traffic lights, pedestrians and intersections.

If the stopping requirements include \textbf{TFL\_RED}, then the stop server has to wait until \textbf{TFL\_GREEN} is issued to satisfy that requirement. When a traffic light state is observed by the perception module, the Event Handler checks whether it is relevant to the current path or not. The light information has to be grounded to an intersection so that the ego car knows where to stop. We use Nearest Neighbor (NN) algorithm to associate detected traffic lights to the intersections.

If the stopping requirements include \textbf{PEDESTRIAN}, then the stop server has to wait until \textbf{PED\_CLEAR} event is received. Both of these events are sent by the Event Handler. Event Handler raises the \textbf{PEDESTRIAN} event if the pedestrian is in a region that is in front of the vehicle, and the pedestrian polygon intersects the road boundaries. The event is accompanied with a stopping point, which is calculated by tracking back a fixed distance from the pedestrian. The stopping point is in turn relayed to the stop server as a requirement. Event Handler can re-issue further events for the same pedestrian, if the person is displaced considerably. If multiple pedestrians are present, only a single event is issued for the pedestrian closest to the ego car.

If the requirements include \textbf{INT}, ego car has to consider other vehicles and safely make the turn. We use a heuristic policy that uses a time to collision (TTC) threshold to decide when to cross. The TTC is defined as follows. Consider an imaginary line starting from the front of the ego car. The TTC with another vehicle is the time for the vehicle to reach that line. We only consider the lowest TTC value among all the cars in the intersection. If the TTC exceeds a threshold for a number of consecutive iterations, then the \textbf{INT\_OK} event is issued. Given data from the vehicle tracker, we compute the TTC by first computing the closing velocity of the crossing traffic (e.g. when the cross traffic is left-right direction, the closing velocity is the vehicle's velocity component in the left-right direction). Then the TTC is calculated as $d_{closing} / v_{closing}$, where $d_{closing}$ is the lateral distance in the left-right direction. Due to the noisy nature of the tracker output, TTC as computed above is noisy and not always monotonically decreasing. We use extrapolation to handle this issue: For a short duration, if the TTC calculated from tracker output is higher than last published TTC, we instead use TTC extrapolated from previous TTC values. Our analysis on TTC performance and comparison with a machine-learning based approach can be found in \cite{bouton2017belief}.

Our analysis on the TTC-based intersection 





\subsection{Trajectory Planning}


\subsubsection{Path Planner}

The path planner uses similar method described in \cite{montemerlo2008junior} to generate several parallel paths around the main center path. After the offset paths are generated and navigation starts, path planner uses pure pursuit algorithm\cite{coulter1992implementation} to constantly correct the ego vehicle motion according to the current pose. It first uses a look ahead distance to query a point on the center path in front of the vehicle. Then it finds the corresponding points on the parallel paths and generates the steering commands toward these points. In this process, we assume constant speed to simplify the calculation. The error caused by this speed inaccuracy is corrected in the next iteration as the planner plans new motion frequently. The computed paths for the center lane and alternatives, along with their corresponding steering commands are sent to the motion evaluator so it can determine the best steering command. Path planner module also keeps track of the vehicle on the global path and publishes the current indexes and look ahead indexes, which are used by behavior layer to trigger events and determine state.

\subsubsection{Motion Evaluator}

The motion evaluator checks all the paths planned by the path planner for obstacles and select the best path to execute. The motion evaluator is agnostic of obstacle types and only handles static obstacles, which are preprocessed as polygons by the Object Aggregator. After receiving the planned paths and obstacle polygons, motion evaluator generates the polygon of the areas driven by the vehicle using the vehicle model for each path and calculate the collision area with all obstacle polygons. The collision area is then used as a score to evaluate a path. Higher score means low collision possibility or no collision and lower score means high collision possibility. The motion evaluator always prefer the center path if there is no obstacle blocking it and prefer the minimal effort to avoid the obstacle. After the evaluation is done, this module also calculates the minimal distance between the vehicle and all the obstacles in front of it as well as the number of collided paths. These information are necessary for the velocity planner described in the next section.

\subsubsection{Velocity Planner}

The velocity planner is responsible for generating a smooth speed profile to achieve a goal position and keep passengers comfortable at the same time. Once the route planner publishes the global path, the velocity planner receives it and generates dynamic speed limit by considering the legal speed limits and road curvature for each waypoint. This module is the last one that runs in trajectory planning because it needs the index of the vehicle in the path and the obstacle information to determine the speed. During the navigation stage, velocity planner monitor the vehicle speed and make sure it is always below the dynamic speed limit. Moreover, the velocity planner plans a speed profile for look ahead distance, obstacle, and sub-goals from behaviour level using S-Curve algorithm\cite{gifre2012adaptive}. The velocity planner plans for all these three targets and chooses the one that is closest to the vehicle or has the lowest profile because the speed profile should always satisfy all these targets and not overshoot.

\subsection{Vehicle Control}

We use a propriety controller for calculation of the steering/throttle and braking commands on the vehicle. The controller expects receiving target trajectory points sampled at 10ms spacing, with 10 points in each communication packet (10ms x 10 points = 100ms of trajectory).  The trajectory planner generates trajectories of the next 3 seconds, at $100ms$ sampling, which is then interpolated to generate the 10 point trajectory packet, sent at $20Hz$. The overlapping half-packet of commanded trajectory is used for redundancy in the communication protocols.


\section{Demonstration}
\label{sec:demo}

To test and highlight the functionality of the designed automated driving system, we devised a demonstration route that encompassed a variety of different scenarios that might be potentially encountered in urban driving.  Our testing was conducted at GoMentum Station in Concord, California around a section of roads with signage, signals and road markings representative of California roadways.  The road speed limits were designated at a maximum of 35mph, though the vehicle calculated maximum speeds based on the posted speed limit as well as dynamic speed limits factoring for maximum lateral acceleration limits in the turns.  For safety, an engineer was present in the driving seat during all testing to monitor the progress of the system and take over control in an event of any system anomalies.  The total length of the demonstration route was 2.5km, encompassing a signalled 4-way intersection, a T-junction, two sweeping curves and 4-way stop.  The scenarios designed to highlight specific aspects and responses of the automated driving system are as follows:

\subsection{Scenario 1: Intersection with Traffic Light}

\vspace{-0.2cm}

\begin{figure}[h!]
\centering     
\subfigure{\label{fig:a}\includegraphics[width=0.235\textwidth]{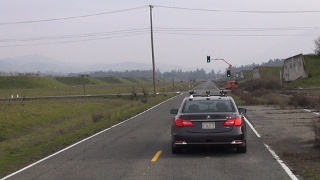}}
\subfigure{\label{fig:b}\includegraphics[width=0.235\textwidth]{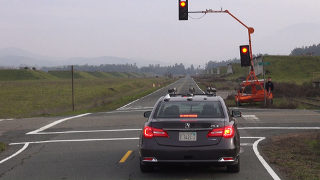}}
\subfigure{\label{fig:a}\includegraphics[width=0.235\textwidth]{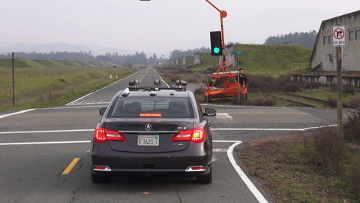}}
\subfigure{\label{fig:b}\includegraphics[width=0.235\textwidth]{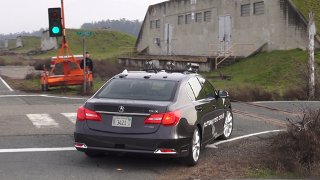}}
\caption{Steps for handling a signalized right turn. a) Green light is detected from a distance of approximately $80m$. Anticipating there is an upcoming right turn, vehicle goes into \textbf{STOP} state. b) The light turns to red, which is correctly recognized, and the AD vehicle comes to a full stop. c) Light turns green, state and the AD vehicle is cleared to continue. Absent cross-traffic, the vehicles resumes to the \textbf{GO} state. d) AD vehicle makes the right turn.}
\vspace{-0.2cm}
\end{figure}

The vehicle approaches a signalled 4-way intersection where the planned route is to make a right turn.  On-board cameras detect the traffic light state, which is initially green on approach to the intersection.  Since there is a rail crossing after the turn, the intersection is signed and designated as ''No Turn On Red''.  As the vehicle approaches, the Enter Intersection event is triggered and the FSM transitions to the STOP state.  As the vehicle detects a transition to a red light, the \textbf{TFL\_RED} event is triggered. The vehicle calculates the stopping point at the intersection where waits for the \textbf{TFL\_GREEN} event.  Once the green light is detected, the \textbf{TFL\_GREEN} event is sent and the vehicle resumes to the \textbf{GO} state.

\subsection{Scenario 2: Turn Right with Cross-Traffic Present}

\vspace{-0.0cm}

\begin{figure}[ht!]
\centering     
\vspace{4pt}
\subfigure{\label{fig:a}\includegraphics[width=0.235\textwidth]{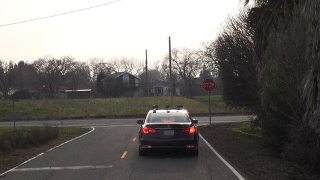}}
\subfigure{\label{fig:b}\includegraphics[width=0.235\textwidth]{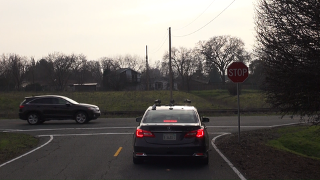}}
\subfigure{\label{fig:a}\includegraphics[width=0.235\textwidth]{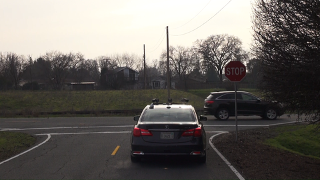}}
\subfigure{\label{fig:b}\includegraphics[width=0.235\textwidth]{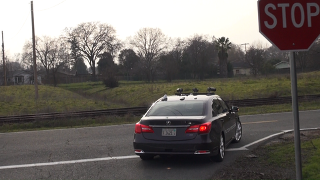}}
\caption{Steps for handling a right turn with cross-traffic present. a) Stop sign is taken into account and the AD vehicle gets into \textbf{STOP} state before the intersection. b) Cross-traffic is detected and the AD vehicle waits until TTC algorithm deems the turn safe. c) Cross-traffic vehicle has passed the intersection, and the AD vehicle resumes to the \textbf{GO} state. d) vehicle makes the right turn.}
\vspace{-4pt}
\vspace{-0.3cm}
\end{figure}

The vehicle approaches a 3-way intersection with a stop sign and comes to a complete stop as the vehicle transitions back to \textbf{STOP}.  As it reaches a stop, oncoming cross-traffic is detected from the left and the ego-vehicle monitors the TTC to determine if there is sufficient time to safely make a right turn ahead of the oncoming vehicle.  If it is judged to not be possible, the vehicle waits until an appropriate gap is present before transitioning back to the \textbf{GO} state.

\subsection{Scenario 3: Navigating a Construction Zone}

\vspace{-0.0cm}

\begin{figure}[h]
\centering     
\subfigure{\label{fig:a}\includegraphics[width=0.235\textwidth]{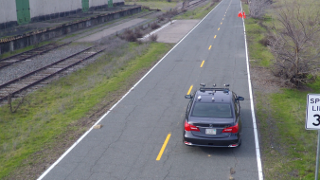}}
\subfigure{\label{fig:b}\includegraphics[width=0.235\textwidth]{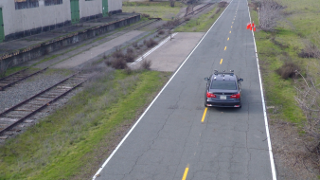}}
\subfigure{\label{fig:a}\includegraphics[width=0.235\textwidth]{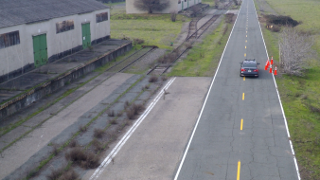}}
\subfigure{\label{fig:b}\includegraphics[width=0.235\textwidth]{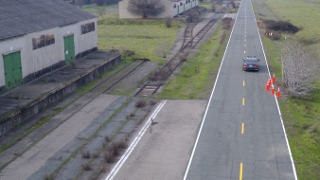}}
\caption{Steps for navigating a construction zone. a) Pylons on the road are recognized using LiDAR, and determines that it is neither a pedestrian nor a vehicle. b) Trajectory planner finds an alternative and obstacle free path and the AD vehicle starts executing the trajectory. c) AD Vehicle drives past obstacles and d) merges back on to the lane.}
\vspace{-0.0cm}
\end{figure}

A portion of the lane is blocked by some construction pylons due to road debris which makes it not possible for the vehicle to strictly follow the lane center without colliding with the cones.  Laser scans detect the existence of the obstacle on the lane/road and the planner calculates to determine if there is enough space to maneuver safely around the obstacle.  If a feasible solution is found, the vehicle smoothly moves laterally from the lane center to avoid the pylons and then returns back to the lane center once past the obstacle.

\subsection{Scenario 4: Stopping for Pedestrian}

\begin{figure}[ht!]
\centering     
\vspace{4pt}
\subfigure{\label{fig:a}\includegraphics[width=0.235\textwidth]{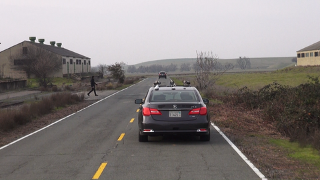}}
\subfigure{\label{fig:b}\includegraphics[width=0.235\textwidth]{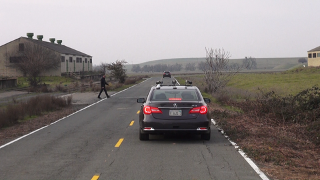}}
\subfigure{\label{fig:a}\includegraphics[width=0.235\textwidth]{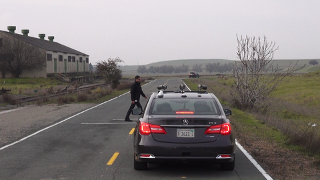}}
\subfigure{\label{fig:b}\includegraphics[width=0.235\textwidth]{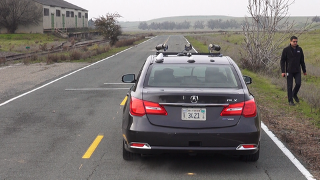}}
\caption{Steps for stopping for a pedestrian. a) A pedestrian is detected that is currently not on the road. b) Pedestrian enters the road, triggering the AD vehicle goes into \textbf{stop} state. c) vehicle remains in \textbf{STOP} state until the pedestrian is cleared. d) vehicle resumes to the \textbf{GO} state after pedestrian leaves the road boundary.}
\vspace{-4pt}
\vspace{-0.4cm}
\end{figure}

\vspace{-0.1cm}

Using a variety of sensing modalities, we were able to demonstrate reactive stopping for both un-occluded and fully occluded pedestrians.  Using the vision-LiDAR sensing modality, the vehicle detects a pedestrian near the roadway and issues a \textbf{PEDESTRIAN} event, resulting transition to the STOP state. A stop point is calculated based on the detected position, allowing the pedestrian to cross the road in front of the vehicle.  With the V2P system, an additional box truck is parked along the edge of the road, completely blocking visibility of the pedestrian walking towards the road.  The vehicle receives the transmitted location of the pedestrian and determines that stopping is necessary.  Once the vehicle comes to a complete stop, the pedestrian appears from behind the truck in front of the vehicle.  In both cases, the vehicle senses that the pedestrian has cleared the roadway (\textbf{PED\_CLEAR}) before transitioning back to the \textbf{GO} state.

\vspace{-0.3cm}

\subsection{Discussion}
During our demonstration days, we completed 44 runs, encompassing 110km of automated driving with 3 cases where the driver had to intervene control of the vehicle.  There were two cases where localization accuracy was temporarily reduced due to poor lane marking recognition coupled with large GPS errors, resulting in inaccuracy of the estimated position of the obstacles detected.  The third case requiring driver intervention occurred when the transmitted position of the pedestrian from the V2P system was inaccurate due to GPS positioning error on the smartphone.  While the vehicle could detect the pedestrian after appearing from behind the truck, the driver overrode the braking response of the automated system on grounds of safety.  During the demonstrations and test runs, the automated driving system was able to perform under both sunny, overcast, light fog and light rain conditions.  With some of the vision-based detection systems, detection ranges were slightly reduced, however the vehicle was still able to respond appropriately to each of the scenarios.

\section{Conclusion and Future Work}
\label{sec:conclusion}

In this paper, we presented a proof-of-concept Acura RLX vehicle, capable of successfully autonomously navigating urban scenarios. These scenarios included intelligently handling of signalized intersections, pedestrians and construction zones. We provided technical descriptions of the underlying modules, including localization, detection/tracking and planning, that work in concert to enable the intelligent behaviors. In a demonstration in GoMentum Station in California, the automated drive vehicle completed 44 runs, with only 3 driver interventions during the $110km$ encompassed. Our demonstration showed that robust and consistent AD behavior in urban scenarios is possible, yet more investigation is necessary for full scale roll-out on public roads.

As future work, we plan to further improve the robustness of the system by addressing the problems that led to driver intervention. We also would like to test our AD capabilities in public roads where the environment is more complex and dynamic.


\vspace{-0.2cm}
\bibliography{refs}

\begin{thebibliography}{10}
\providecommand{\url}[1]{#1}
\csname url@samestyle\endcsname
\providecommand{\newblock}{\relax}
\providecommand{\bibinfo}[2]{#2}
\providecommand{\BIBentrySTDinterwordspacing}{\spaceskip=0pt\relax}
\providecommand{\BIBentryALTinterwordstretchfactor}{4}
\providecommand{\BIBentryALTinterwordspacing}{\spaceskip=\fontdimen2\font plus
\BIBentryALTinterwordstretchfactor\fontdimen3\font minus
  \fontdimen4\font\relax}
\providecommand{\BIBforeignlanguage}[2]{{%
\expandafter\ifx\csname l@#1\endcsname\relax
\typeout{** WARNING: IEEEtran.bst: No hyphenation pattern has been}%
\typeout{** loaded for the language `#1'. Using the pattern for}%
\typeout{** the default language instead.}%
\else
\language=\csname l@#1\endcsname
\fi
#2}}
\providecommand{\BIBdecl}{\relax}
\BIBdecl

\bibitem{nhtsa}
\BIBentryALTinterwordspacing
NHTSA, \emph{Quick Facts 2015}, 2015 (accessed January 29, 2017). [Online].
  Available:
  \url{https://crashstats.nhtsa.dot.gov/Api/Public/ViewPublication/812348}
\BIBentrySTDinterwordspacing

\bibitem{urmson2008autonomous}
C.~Urmson, J.~Anhalt, D.~Bagnell, C.~Baker, R.~Bittner, M.~Clark, J.~Dolan,
  D.~Duggins, T.~Galatali, C.~Geyer \emph{et~al.}, ``Autonomous driving in
  urban environments: Boss and the urban challenge,'' \emph{Journal of Field
  Robotics}, vol.~25, no.~8, pp. 425--466, 2008.

\bibitem{montemerlo2008junior}
M.~Montemerlo, J.~Becker, S.~Bhat, H.~Dahlkamp, D.~Dolgov, S.~Ettinger,
  D.~Haehnel, T.~Hilden, G.~Hoffmann, B.~Huhnke \emph{et~al.}, ``Junior: The
  stanford entry in the urban challenge,'' \emph{Journal of field Robotics},
  vol.~25, no.~9, pp. 569--597, 2008.

\bibitem{ziegler2014making}
J.~Ziegler, P.~Bender, M.~Schreiber, H.~Lategahn, T.~Strauss, C.~Stiller,
  T.~Dang, U.~Franke, N.~Appenrodt, C.~G. Keller \emph{et~al.}, ``Making bertha
  drive—an autonomous journey on a historic route,'' \emph{IEEE Intelligent
  Transportation Systems Magazine}, vol.~6, no.~2, pp. 8--20, 2014.

\bibitem{cui_ekf_urban}
Y.~Cui and S.~Sam~Ge, ``Autonomous vehicle positioning with gps in urban canyon
  environments,'' \emph{IEEE Transactions on Robotics and Automation}, vol.~19,
  no.~1, pp. 15--25, 2003.

\bibitem{vdw_ukf}
R.~van~der Merwe and A.~Wan, Eric, ``Sigma-point kalman filters for integrated
  navigation,'' in \emph{Proceedings of the 60th Annual Meeting of The
  Institute of Navigation (2004)}.\hskip 1em plus 0.5em minus 0.4em\relax ION,
  2004, pp. 641--654.

\bibitem{tao_localization}
Z.~Tao, P.~Bonnifait, V.~Frémont, and J.~Ibañez-Guzman, ``Lane marking aided
  vehicle localization,'' in \emph{16th International IEEE Conference on
  Intelligent Transportation Systems (ITSC 2013)}.\hskip 1em plus 0.5em minus
  0.4em\relax IEEE, 2013, pp. 1509--1515.

\bibitem{gruyer_localization}
D.~Gruyer, R.~Belaroussi, and M.~Revilloud, ``Map-aided localization with
  lateral perception,'' in \emph{2014 IEEE Intelligent Vehicles Symposium
  Proceedings}.\hskip 1em plus 0.5em minus 0.4em\relax IEEE, 2014, pp.
  674--680.

\bibitem{jo_localization}
K.~Jo, K.~Chu, and M.~Sunwoo, ``Gps-bias correction for precise localization of
  autonomous vehicles,'' in \emph{2013 IEEE Intelligent Vehicles Symposium
  (IV)}.\hskip 1em plus 0.5em minus 0.4em\relax IEEE, 2013, pp. 636--641.

\bibitem{heisele2017system}
B.~Heisele and A.~Ayvaci, ``System and method for providing laser camera fusion
  for identifying and tracking a traffic participant,'' U.S. Patent
  14/876\,907, Apr. 7, 2017.

\bibitem{knill2016direct}
C.~Knill, A.~Scheel, and K.~Dietmayer, ``A direct scattering model for tracking
  vehicles with high-resolution radars,'' in \emph{Intelligent Vehicles
  Symposium (IV), 2016 IEEE}.\hskip 1em plus 0.5em minus 0.4em\relax IEEE,
  2016, pp. 298--303.

\bibitem{dao2007markov}
T.-S. Dao, K.~Y.~K. Leung, C.~M. Clark, and J.~P. Huissoon, ``Markov-based lane
  positioning using intervehicle communication,'' \emph{IEEE Transactions on
  Intelligent Transportation Systems}, vol.~8, no.~4, pp. 641--650, 2007.

\bibitem{bouton2017belief}
M.~Bouton, A.~Cosgun, and M.~J. Kochenderfer, ``Belief state planning for
  navigating urban intersections,'' \emph{IEEE Intelligent Vehicles Symposium
  (IV)}, 2017.

\bibitem{coulter1992implementation}
R.~C. Coulter, ``Implementation of the pure pursuit path tracking algorithm,''
  DTIC Document, Tech. Rep., 1992.

\bibitem{gifre2012adaptive}
C.~Gifre~Oliveras, ``Adaptive motion planning for a mobile robot,'' Master's
  thesis, Universitat Polit{\`e}cnica de Catalunya, 2012.

\end{thebibliography}
\bibliographystyle{IEEEtran}
\end{document}